\documentclass{IEEEcsmag}

\usepackage[colorlinks,urlcolor=blue,linkcolor=blue,citecolor=blue]{hyperref}
\expandafter\def\expandafter\UrlBreaks\expandafter{\UrlBreaks\do\/\do\*\do\-\do\~\do\'\do\"\do\-}
\usepackage{upmath,color}
\usepackage{booktabs} 

\jvol{58}
\jnum{11}
\paper{8}
\jmonth{November}
\jname{IEEE Computer}
\jtitle{Our Cars Can Talk: How IoT Brings AI to Vehicles}
\pubyear{2025}

\begin{document}

\sptitle{Humanity and Computing, IEEE Computer}

\title{Our Cars Can Talk: How IoT Brings AI to Vehicles}

\author{Amod K. Agrawal}

\affil{Amazon Lab126, Sunnyvale, California, 94089, USA}

\markboth{Humanity and Computing}{Humanity and Computing}

\begin{abstract}
Bringing AI to vehicles and enabling them as sensing platforms is key to transforming maintenance from reactive to proactive. Now is the time to integrate AI copilots that speak both languages: machine and driver. This article offers a conceptual and technical perspective intended to spark interdisciplinary dialogue and guide future research and development in intelligent vehicle systems, predictive maintenance, and AI-powered user interaction.
\end{abstract}

\maketitle


\chapteri{W}hile personal vehicles have essentially become drivable computers, the consumer experience of vehicle maintenance remains outdated and hasn’t changed since the early 2000s. Vehicle maintenance remains largely reactive to this day, often triggered by the dreaded check engine light, sometimes at the worst possible time: in the middle of a busy week, or right before a road trip. However, today’s vehicles are equipped with a dense network of sensors that can monitor nearly every aspect of performance in real time. 

\section{Automotive as an IoT sensor}
The Society of Automotive Engineers (SAE) standardized the OBD-II interface in mid-1990s to access the vehicle’s Engine Control Unit (ECU) and its data streams, which can be used to monitor faults and perform diagnostics.$^{\cite{obd}}$ It allows real-time readouts of key parameters and metrics for engine performance, fuel system, fuel trim, emissions, air system, battery and voltage, and other sensors throughout the vehicle. 
The automotive industry traditionally moves slowly with new software paradigms, whereas the AI world is accelerating rapidly. As AI becomes increasingly embedded in our daily lives, through personal devices and smart homes, it should naturally grow to integrate seamlessly with our vehicles as well. Vehicles offer a rich sensing environment that can provide valuable context to personalize the AI agents further. Similarly, bringing AI into the vehicle ecosystem through natural language interfaces and real-time data mining models opens up exciting possibilities. By fusing time series sensor data from the OBD interface with intelligent assistants on-board, we can make maintenance proactive and enhance vehicle interaction to be contextual and deeply personalized. This two-way fusion of information can significantly improve the vehicle’s ownership and maintenance experience, while also extending the conversational voice-based interactions to the cabin.

\begin{figure*}[t]
  \centering
  \includegraphics[width=\textwidth]{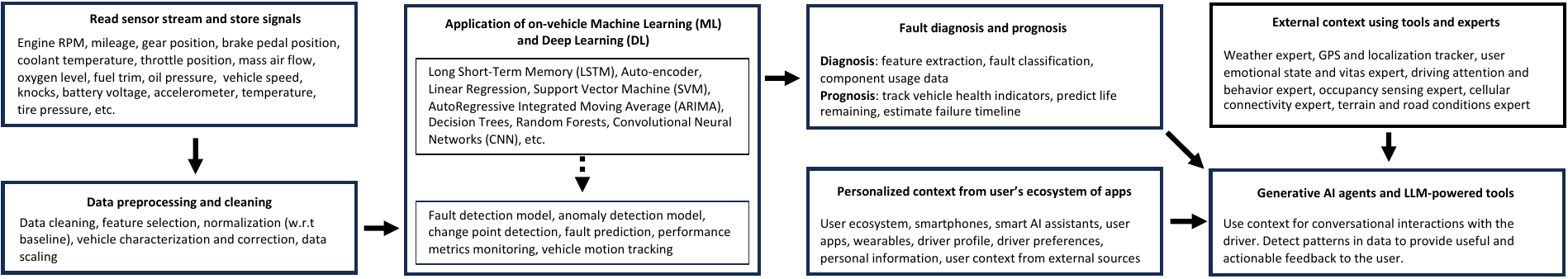}
  \caption{Pipeline for predictive maintenance and AI-driven vehicle interaction. The system begins by collecting sensor data from the vehicle’s OBD and embedded systems, capturing parameters such as RPM, throttle position, fuel trim, and tire pressure. This data is preprocessed through cleaning, normalization, and correction for vehicle-specific baselines. Machine learning and deep learning models are then applied on the vehicle to detect anomalies, predict faults, and track performance. The pipeline also incorporates contextual signals from the user’s ecosystem (e.g., smartphones, wearables, preferences) and external sources (e.g., GPS, road conditions, emotional state). Generative AI agents use this fused context to deliver personalized, conversational feedback and proactive maintenance recommendations.}
  \label{fig:pipeline}
\end{figure*}

\section{Predictive maintenance: IoT as an Enabler for AI}
Predictive maintenance has been a focus in industrial IoT and fleet logistics for some time,$^{\cite{predictivemaintenance, fleetmanagement}}$ but its integration with edge AI and large language models (LLMs) in consumer vehicles remains largely unexplored. Presently, most connected maintenance services in modern vehicles simply read OBD data and notify users when a fault code is triggered. Google and Volkswagen recently announced a partnership to use Gemini models to help drivers understand vehicle status and assist in diagnostics.$^{\cite{googlevw}}$ While this is a promising start, it leaves untapped potential in combining real-time sensor streams with on-vehicle anomaly detection models, and using failure metrics as contextual signals for AI agents. 

Machine Learning models such as LSTMs and ARIMA can identify early signs of performance degradation by detecting anomalies and deviations in vehicle performance metrics.$^{\cite{mlpredictive}}$ Recurring patterns in historical data, particularly diagnostic trouble codes (DTCs) and long-term sensor time series, can be used to train failure classification models. Further, statistical and Deep Learning models can be trained to predict faults based on sensor anomalies and detecting unusual patterns in vehicle behavior over time. These models can also adapt to external factors such as driving behavior, trip length, terrain, and even weather conditions. This is just the starting point as there’s a broader opportunity to leverage pattern recognition algorithms across multiple modalities of data which can include—driver profile and preferences, driving behavior, vehicle performance metrics, and external context from other sensing sources. Personalized AI agents (such as Alexa, Siri, etc.) which are integrated across the user’s ecosystem already have access to rich contextual data such as home and work locations, routing history via Google or Apple Maps, trip details and frequency, driving conditions (e.g., highways vs. city streets), preferred gas stations or service centers, and even vehicle’s energy type (gasoline, hybrid, or electric). When paired with real-time vehicle performance data, this not only enables holistic vehicle health tracking but also can create meaningful feedback loops for the consumer. For instance, a highway-heavy driver may require different service intervals than someone commuting in city traffic. Vehicles exposed to extreme heat or cold will age differently, and predictive models could account for that. Frequent hard braking or long idle times can shape predictions around brake wear, battery health, or engine stress. By interpreting trends across multiple signals, AI can identify subtle patterns, such as wheel RPM discrepancies suggesting end-of-life for a bearing, or declines in fuel efficiency that may indicate clogged injectors or filter issues. Predictive maintenance improves when it learns from a wide range of signals, allowing it to adapt to each driver and their surroundings.

\section{AI Assistants as Copilots in the Cabin}
Most modern vehicles now offer smart assistants, but their functionality is still limited to voice interaction, navigation, and entertainment. A few advanced systems go further, incorporating ambient RF-based sensing to detect occupancy and presence,$^{\cite{occupanydetection}}$ measure emotional states through breathing and heart rate,$^{\cite{breathingrate, vitalmonitoring}}$ and monitor driver attention using eye tracking and reaction times.

Fusion of these sensing technologies with AI-powered intelligence showcases a shift from reactive systems to context-aware agents which can enable innovative interactions with the driver:

\begin{itemize}
    \item \textit{“You're braking harder than usual this week. That could wear your brake pads faster. Next Saturday seems open on your calendar, should I schedule a check at your preferred service center?”}
    \item \textit{“Your tire pressure drops every time temperature falls below $28^\circ F$. Next week, temperatures are expected to go as low as $20^\circ F$, should I remind you to make a stop at the nearest gas station?”}
    \item \textit{“I’ve detected two engine misfire events this week. Should I book a service appointment at your usual garage this Sunday?”}
    \item \textit{“You’ve consumed more fuel than usual this week. Your vitals indicate that you might be tired. Should I add a stop at the gas station on way to work tomorrow?”}
\end{itemize}	
Figure \ref{fig:pipeline} showcases the proposed end-to-end pipeline of information flow from the different sensing modalities and their fusion using AI methods to enable personalized and conversational feedback for the users.

\section{Scalability and Privacy}

While the OBD interface and its protocols have been standardized, baseline values and signal characteristics can vary widely across OEMs, models, and even firmware versions. Therefore, training Machine Learning models that can scale and generalize across heterogeneous vehicle types and configurations can be a complex problem. Federated Learning (FL) offers a scalable, privacy-preserving yet practical solution by allowing training of models locally on the vehicles while sharing only the updated weights or gradients with a central aggregator in the cloud. However, FL faces challenges such as non-IID data (each vehicle collects distinct, often skewed data distributions), constrained on-board resources, and intermittent connectivity.$^{\cite{heterogenousvehicles}}$ These factors can cause local models to diverge significantly from the global objective. Specialized algorithms like Federated Average, Federated Proximal, and Scaffold are designed to improve convergence and model consistency across diverse and decentralized fleets.$^{\cite{federatedlearning}}$ On-device fine-tuning enables models to adapt to different regions, terrains, driver demographics, and vehicle configurations. Techniques like differential privacy and secure aggregation can prevent leakage of sensitive information and ensure user-privacy.

\section{Beyond Maintenance: Enabling Ecosystem of Smart Environments}
The fusion of AI and vehicle data unlocks new use cases that may extend beyond predictive maintenance. For example, AI-powered driving coaches can provide feedback to teens or new drivers based on their behavior in real-time. Fleet operators can monitor vehicle health across large deployments, optimize service schedules, and efficiently manage inventory. Insurance providers may integrate with systems to reward drivers with safe driving habits and consistent service records. Smart vehicle environments have the potential to integrate seamlessly with smartphones, wearables, and home automation systems to deliver a truly coherent and connected experience for consumers. In this ecosystem, personalized AI is as effective as the quality and depth of data it can access, and personal vehicles play a critical role in enriching its context.

\def\refname{REFERENCES}

\begin{IEEEbiography}{Amod K. Agrawal} is an Applied Scientist at \linebreak Amazon Lab126 at Sunnyvale, California, 94089. His research interests include mobile computing, wireless sensing, and ambient computing on the edge. Agrawal received his M.S. in Computer Science from the University of Illinois, Urbana-Champaign. He is a Senior Member of the IEEE, and a contributing member of the IEEE Computer Society.\linebreak Contact him at amoagraw@amazon.com.\vspace*{8pt}
\end{IEEEbiography}
\end{document}